\documentclass[10pt,journal,compsoc, twoside]{IEEEtran}
\usepackage{graphicx}
\usepackage{xcolor}
\usepackage{amstext}
\usepackage{amsmath}
\usepackage{amssymb}
\usepackage{float}
\usepackage{booktabs}
\usepackage{tikz}

\usepackage[flushleft]{threeparttable}
\usepackage{array, multirow, multicol}
\DeclareMathOperator{\tr}{tr}

\newcommand\norm[1]{\left\lVert#1\right\rVert}

\usepackage{tabularx,ragged2e}
\usepackage{algorithm}
\usepackage{algpseudocode}
\usepackage{caption}

\tikzset{%
	block/.style    = {draw, thick, rectangle, minimum height = 3em,
		minimum width = 3em},
	sum/.style      = {draw, circle, node distance = 2cm}, 
	input/.style    = {coordinate}, 
	output/.style   = {coordinate} 
    >=stealth',
    punkt/.style={
           rectangle,
           rounded corners,
           draw=black, very thick,
           text width=6.5em,
           minimum height=2em,
           text centered},
    pil/.style={
           ->,
           thick,
           shorten <=2pt,
           shorten >=2pt,}
}

\tikzstyle{branch}=[fill,shape=circle,minimum size=3pt,inner sep=0pt]
\usetikzlibrary{shapes,arrows}

\algnewcommand\algorithmicinput{\textbf{Input:}}
\algnewcommand\Input{\item[\algorithmicinput]}
\algnewcommand\algorithmicoutput{\textbf{Output:}}
\algnewcommand\Output{\item[\algorithmicoutput]}

\ifCLASSOPTIONcompsoc
\usepackage[nocompress]{cite}
\else
\usepackage{cite}
\fi

\ifCLASSINFOpdf
\else
\fi
\begin{document}

\title{Learnable Graph-regularization for Matrix Decomposition}
	
\author{Penglong~Zhai~and~Shihua~Zhang*
\IEEEcompsocitemizethanks{\IEEEcompsocthanksitem Penglong Zhai and Shihua Zhang are with the NCMIS, CEMS, RCSDS, Academy of Mathematics and Systems Science, Chinese Academy of Sciences, Beijing 100190, School of Mathematical Sciences, University of Chinese Academy of Sciences, Beijing 100049, and Center for Excellence in Animal Evolution and Genetics, Chinese Academy of Sciences, Kunming 650223, China. \protect\\ *To whom correspondence should be addressed. Email: zsh@amss.ac.cn.}}

\markboth{ }%
{ }

\IEEEtitleabstractindextext{%
\begin{abstract}
Low-rank approximation models of data matrices have become important machine learning and data mining tools in many fields including computer vision, text mining, bioinformatics and many others. They allow for embedding high-dimensional data into low-dimensional spaces, which mitigates the effects of noise and uncovers latent relations. In order to make the learned representations inherit the structures in the original data, graph-regularization terms are often added to the loss function. However, the prior graph construction often fails to reflect the true network connectivity and the intrinsic relationships. In addition, many graph-regularized methods fail to take the dual spaces into account. Probabilistic models are often used to model the distribution of the representations, but most of previous methods often assume that the hidden variables are independent and identically distributed for simplicity. To this end, we propose a learnable graph-regularization model for matrix decomposition (LGMD), which builds a bridge between graph-regularized methods and probabilistic matrix decomposition models. LGMD learns two graphical structures (i.e., two precision matrices) in real-time in an iterative manner via sparse precision matrix estimation and is more robust to noise and missing entries. Extensive numerical results and comparison with competing methods demonstrate its effectiveness.
\end{abstract}
		
}

\maketitle

\IEEEdisplaynontitleabstractindextext
	
\IEEEpeerreviewmaketitle

\IEEEraisesectionheading{\section{Introduction}\label{sec:introduction}}
\IEEEPARstart{M}any data of diverse fields are redundant and noisy. Extracting useful information from such primitive data directly is often infeasible. Therefore, how to discover compact and meaningful representations of high-dimensional data is a fundamental problem. Researchers have developed many powerful methods to address this issue from different views. Matrix decomposition approaches have been successfully applied to various fields in the past two decades \cite{zhang2012discovery,zhang2019learning,zhang2019bayesian}. A classical low-rank matrix decomposition is achieved by minimizing a loss or error function between an observed measurement matrix and a bilinear factorization. Specifically, for a given data matrix $Y \in \mathcal{R}^{n \times p}$, its low-rank factorization model is written as follows
\begin{equation}
		Y = XW^{T}+E
\end{equation}
where $X\in \mathcal{R}^{n\times k}$, $W\in \mathcal{R}^{p\times k}$ and $E$ is the noise matrix. There are mainly two ways on how to obtain better representations. One is to model the noise matrix $E$ of the original data and the other is to model the subspace structures of the representations or factor matrices $X$ and $W$.

\par Principal component analysis (PCA) is one of the most fundamental and widely used matrix decomposition method, which has become a standard tool for dimension reduction and feature extraction in various fields such as signal processing \cite{moore1981principal}, human face recognition \cite{hancock1996face}, gene expression analysis \cite{hastie2000gene}. PCA obtains a low-dimensional representation for high-dimensional data in a $l_2$-norm sense. However, it is known that the conventional PCA is sensitive to gross errors. To remove the effect of sparse gross errors, robust PCA (RPCA) \cite{wright2009robust, xu2010robust, candes2011robust} has been proposed. RPCA seeks to decompose the data matrix as a superposition of a low-rank matrix with a sparse matrix. The sparse matrix captures the gross errors and enables RPCA to recover the underlying low-rank representation of the primitive data with vast applications \cite{candes2011robust, peng2012rasl, liu2013robust, shahid2015robust}.
	
\par Tipping and Bishop \cite{tipping1999probabilistic} originally presented that PCA can be derived from a Gaussian latent variable model (named as the probabilistic PCA, PPCA), which became the beginning of the probabilistic decomposition models. PPCA assumes that the underlying distribution of noise is the independent identical distribution (IID) of Gaussian. Its maximum likelihood estimation leads to a PCA solution. This probabilistic framework allows various extensions of PCA and matrix decomposition, and becomes a natural way to model different types of noise or error of data. Different prior distributions have been introduced into this framework. For example, Zhao and Jiang \cite{zhao2006probabilistic} used the multivariate Student-t distribution to model the noise (tPPCA). Wang  \textit{et al.} \cite{wang2012probabilistic} introduced the Laplace distribution to model the noise (RPMF), which is more robust to outliers. Li and Tao \cite{li2010simple} employed the exponential family distributions to handle general types of noise. Zhao \textit{et al.} \cite{zhao2014robust} introduced the mixture of Gaussian to model the complex noise. However, most of those variants assume the underlying distribution of noise is IID, which rarely holds in reality. More recently, Zhang \textit{et al.} \cite{zhang2019matrix} proposed a novel probabilistic model, which places the matrix normal prior to the noise to achieve graphical noise modeling.
	
\par On the other hand, a number of studies have been developed to model the representation structures by applying diverse regularization terms to the loss functions \cite{min2018group},\cite{zhang2011novel},\cite{zhang2019general}. A common assumption is that if two data points are close in the original space, their representations should be close to each other too. To preserve the local geometrical structure in representation, graph regularizers have been imposed. Gao \textit{et al.} \cite{gao2010local} proposed a sparse coding method to exploit the dependence among the feature space by constructing a Laplacian matrix. Zheng \textit{et al.} \cite{zheng2011graph} also proposed a graph-regularized sparse coding model to learn the sparse representations that explicitly takes into account the local manifold structure of the data. More recently, Yin \textit{et al.} \cite{yin2015dual} proposed a low-rank representation method to consider the geometrical structure in both feature and sample spaces. As many measurements in experiments are naturally non-negative, some non-negative matrix factorization variants with graph-regularization have also been developed \cite{cai2011graph, zhang2011novel}. Nevertheless, the graph-construction is far rather arbitrary. It may not accurately reflect the true network connectivity and the intrinsic relationships between data entities (e.g., negatively conditional dependency). Probabilistic models on matrix decomposition have also been tried to model structure of the representations by placing different priors on $X$ and/or $W$ \cite{mnih2008probabilistic},\cite{lim2007variational},\cite{salakhutdinov2008bayesian}. However, in previous probabilistic models, both sample and feature spaces of hidden representations are always assumed to be IID, which rarely holds in real-world scenarios.
	
\par To this end, we propose a matrix decomposition model with learnable graph-regularization (LGMD) to obtain the low-rank representation and the structure of the underlying latent variables simultaneously. The key idea is to model the latent variables as matrix normal distributions. It enables us to explore the structures of the latent variables in both sample and feature spaces in real-time and better estimate graphs with the hidden information extracted from the data. The parameters of LGMD can be estimated with a hybrid algorithm of the alternative least square (ALS) and sparse precision matrix estimation methods under the block coordinate descent framework. Extensive experiments on various data show the effectiveness of LGMD by considering the structure of the underlying sample and feature manifolds. In short, LGMD can obtain a better low-rank representation and a better restoration of the original data with the learned sample and feature structures.

\section{Related Work}
\subsection{Probabilistic Models on Matrix Decomposition}
\par Low-rank matrix decomposition is a large class of methods to achieve the low-rank approximation of a given data matrix. The conventional matrix decomposition models are based on the assumption that the data matrices are contaminated stochastically with diverse types of noises and the low-rank matrices are deterministic with unknown parameters. Thus, the point estimations of low-rank components can be obtained by maximum likelihood estimation or maximum a posteriori. A prominent advantage of the aforementioned point estimation methods is that they are simple and easy to implement. However, we can not obtain the probability distributions of the low-rank matrices that are pre-requisite in exploring the generative models. In the past two decades, a variety of probabilistic models of low-rank matrix decomposition have been developed. The most significant difference between low-rank matrix decomposition methods and their corresponding probabilistic models is that the latter treat the low-rank components as random variables. These probabilistic models have been widely applied onto the fields of signal and image processing, computer vision, bioinformatics and so on.

\par Tipping and Bishop \cite{tipping1999probabilistic} originally presented PPCA by assuming the latent variables following the unit isotropic Gaussian distribution. Specifically, the generative model of PPCA is
\begin{equation}
	Y=XW^T+E \label{eq:gen_pca}
\end{equation}
where $X\in R^{n\times k }$, $W\in R^{p \times k}$ and $E_{ij} \overset{IID}{\sim} \mathcal{N}(0, \sigma^2)$. The negative log likelihood is as follows
\begin{equation}
	\frac{1}{2\sigma^2} \norm{Y- XW^T}_F^2 + mn \log \sigma \label{eq:obj_ppca}
\end{equation}
where $\sigma$ is the standard deviation of the noise and $\norm{\cdot}_F$ denotes the Frobenius norm.
If we treat $\sigma$ as a constant, the objective function Eq. (\ref{eq:obj_ppca}) is equivalent to the minimization of the reconstruction error in PCA. PPCA further places the standard Gaussian $\mathcal{N}(0, I)$ prior on each column of $X$ and derive the maximum likelihood estimation of $W$ and $\sigma$.
	
\par The goal of PPCA is not to give better results than PCA, but to permit a broad range of future extensions by facilitating various probabilistic techniques and introducing different assumptions of prior distributions. For example, Bishop \cite{bishop1999variational} developed a variational formulation of Bayesian PCA, which can automatically determine the number of retained principal components. Zhao and Jiang \cite{zhao2006probabilistic} proposed tPPCA which assumes that data are sampled from the multivariate Student-t distribution. Probabilistic matrix factorization (PMF) \cite{mnih2008probabilistic} assumes that $X_{ik}$ are independent and identically distributed and so are $W^{T}_{kj}$, and Variational Bayesian PMF \cite{lim2007variational} assumes the entries from different columns of $X$ or $W$ have different variances. Bayesian PMF \cite{salakhutdinov2008bayesian} further generalizes PMF by assuming that columns of $X$ or $W$ are multivariate Gaussian distributed and the rows are independent and identically distributed. Among all of these probabilistic matrix factorization models, PMF can be a typical case. Similar to PPCA, it follows the same generative model, while places isotropic Gaussian priors on $X$, $W$ and the noise
\begin{equation*}
E_{ij} \overset{IID}{\sim} \mathcal{N}(0, \sigma^2), X_{ik} \overset{IID}{\sim} \mathcal{N}(0, \sigma_{X}^2), W^{T}_{kj} \overset{IID}{\sim} \mathcal{N}(0, \sigma_{W}^2)
\end{equation*}
Its negative log-posterior over the observed is
\begin{equation}
	\frac{1}{2\sigma^2} \norm{Y- XW^T}_F^2 + \frac{\lambda_X}{2}\norm{X}^{2}_F+\frac{\lambda_W}{2}\norm{W}_F^2 \label{eq:obj_pmf}
\end{equation}
where $\lambda_{X} = \sigma^{2}/\sigma_{X}^{2}$, $\lambda_W=\sigma^{2}/\sigma_{W}^{2}$.

\begin{figure*}
		\centering
		\includegraphics[width=0.84\textwidth]{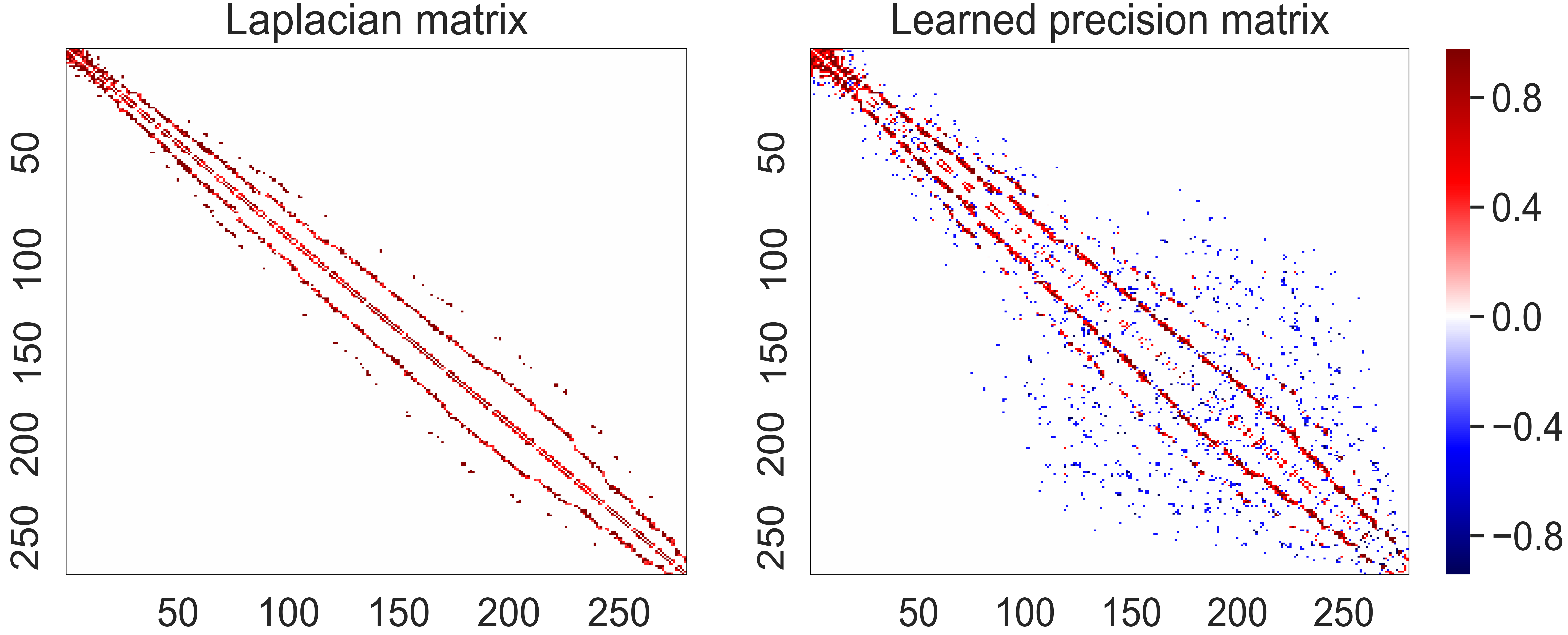}
		\caption{Illustration of the Laplacian and precision matrices based on a data with 250 samples \cite{srivastava2017large}.}
		\label{fig:compare_LP}
\end{figure*}

\par Most of these methods assume that the underlying distributions of both sample and feature spaces are independent and identically distributed for simplicity. However, the structures of the sample manifold and feature manifold might be complicated and nonlinear, which are often ignored in previous probabilistic models. To address this challenge, Zhang \textit{et al.} \cite{zhang2019matrix} recently proposed a novel probabilistic model on matrix decomposition by placing the matrix normal prior on the noise to explore the structures of sample and feature spaces. While their focus is on graphical noise modeling but not about the latent variables.
	
\subsection{Graph-regularized Matrix Decomposition}
\par Graph-regularized matrix decomposition (GRMD) methods obtain the low-rank representation of the primitive data and preserve the local geometrical structure by graph regularizers to some extent. For example, Zheng \textit{et al.}  \cite{zheng2011graph} proposed a graph-regularized sparse coding method for image presentation. It has the following form
\begin{equation}
	\min \norm{Y - XW^T}_F^2 + \eta \tr(X^T L  X)  + \rho \norm{X}_1
\end{equation}
where $Y \in R^{n \times p}$ is the data matrix, $\eta > 0$, $\rho > 0$, $X$ is the low-dimensional representation of images, $L\in R^{n \times n }$ is the Laplacian matrix, and the $l_1$-norm regularizer encourages $X$ to be sparse. One can construct a binary graph matrix $G$ by the $k$-NN algorithm
\begin{equation}
		G_{ij}=
		\begin{cases}
			1,& \text{if } x_i \text{ is a neighbor of } x_j \\
			0,              & \text{otherwise}
		\end{cases}
\end{equation}
The Laplacian matrix is defined as $L  =  D - G$, where $D = \mbox{diag}(d_1,d_2,\cdots,d_n)$ ($d_j=\sum_{i=1}^n G_{ij}$) is the degree matrix. Then the graph regularizer can be written as
\begin{equation}
		\tr(X^T L  X) = \frac{1}{2} \sum_{ij} (x_i - x_j)^2 G_{ij}
\end{equation}
and it encourages neighbors in the original space to be neighbors in the sparse representation $X$.

\par Recent studies have shown that not only the observed data are found to lie on a nonlinear low-dimensional manifold, namely sample manifold, but also the features lie on a manifold, namely feature manifold. For example, Yankelevsky and Elad \cite{yankelevsky2016dual} proposed a low-rank representation method to consider the geometrical structure in both feature and sample spaces in the following manner
\begin{align}
		\begin{split}
			\min_{X, W} & \norm{Y - XW^T}_F^2 + \eta_1 \tr(X^TL X) + \eta_2 \tr(W^T L_c W) \\
			\text{s.t.} & \norm{x_i}_0 \leq T \quad \forall i
		\end{split}
\end{align}
where $L$ and $L_c$ are the Laplacian matrices of sample and feature spaces respectively. $T$ is a parameter to control the sparsity of $x_i$. However, the Laplacian matrix prefers to consider the local geometry while the precision matrix not only captures the neighbor relationships between the variables but also recovers relationships between variables that are negatively condition-dependent (Fig. \ref{fig:compare_LP}) \cite{srivastava2017large}. In addition, conventional graph-regularization is derived based on the original data directly, which doesn't consider the intrinsic relationships among latent variables.

\subsection{Sparse Precision Matrix Estimation} \label{sec:sparse_precision}
Precision matrix (i.e., the inverse of the covariance matrix) reveals the conditional correlations between pairs of variables. How to estimate a large precision matrix is a fundamental issue in modern multivariate analysis. Formally, suppose we have $n$ multivariate normal observations of dimension $p$ with covariance $\Sigma$. Let $\Theta = \Sigma^{-1}$ be the precision matrix and $S$ be the empirical covariance matrix, then the problem of precision matrix estimation is to maximize the log-likelihood
\begin{equation}
		\ln |\Theta| -\tr(S \Theta) \label{eq:precision_ll}
\end{equation}
where $\Theta$ is a positive definite matrix, and $|M|$ is the determinant of a matrix $M$.

\begin{figure*}
	\begin{tikzpicture}[auto, thick, node distance=2cm, >=triangle 45,scale=0.95]
		\draw node at (3,6.8) [sum, name=A] {\large $A_{}$};
		\draw node at (11,6.8) [sum, name=B] {\large $B_{}$};
		\draw node at (7,5) [sum,fill=blue!10, name=Y] {$Y_{ij}$};
		\draw node at (5,5) [sum, name=X] {$X_{i}$};
		\draw node at (9,5) [sum, name=W] {$W_{j}$};
		
		\node at (7,3) [below=5mm, right=-2.2mm,name=sigma] {\large $\sigma$};
		\node at (7,3)[below=3mm,right=-1.9mm]{\textbullet};
		\node at (3,5) [below=0mm, right=-3mm,name=sigmax] {\large $\sigma_{X}$};
		\node at (3,5)[below=0.2mm, right=3mm]{\textbullet};
		\node at (11,5) [below=0mm, right=-4mm,name=sigmaw] {\large $\sigma_W$};
		\node at (11,5)[below=0.2mm, right=-7mm]{\textbullet};
		\node at (3,9) [below=5mm, right=-3mm,name=lambda1] {\large $\eta_A$};
		\node at (3,9)[below=7.5mm,right=-1.8mm]{\textbullet};
		\node at (11,9) [below=5mm, right=-3mm,name=lambda2]{\large $\eta_B$};
		\node at (11,9)[below=7.5mm,right=-1.8mm]{\textbullet};
		
		\path[bend left=10]
        (X) edge[->] (A)
        (A) edge[->] (X);

		\path[bend left=10]
        (W) edge[->] (B)
        (B) edge[->] (W);
		\draw[->] (sigmax) -- node {}(X);	
		\draw[->] (sigmaw) -- node {}(W);	
		\draw[->] (X) -- node {}(Y);
		\draw[->] (W) -- node {}(Y);
		\draw[->] (sigma) -- node {}(Y);
		\draw[->] (lambda1) -- node {}(A);
		\draw[->] (lambda2) -- node {}(B);
		\draw [color=gray,thick](4,3) rectangle (8,7);
		\draw [color=gray,thick](6,3.5) rectangle (10,7.5);
		\node at (4,3) [above=5mm, right=0mm] {$i \in [0,n]$};\
		\node at (10,3.5) [above=5mm, right=-15mm] {$j \in [0,p]$};
		\node at (1.5,8.3) [below=0mm, right=0mm] {\Large (b)};
				
		
		\draw node at (-3,5) [sum,fill=blue!10, name=Y1] {$Y_{ij}$};
		\draw node at (-5,5) [sum, name=X1] {$X_{i}$};
		\draw node at (-1,5) [sum, name=W1] {$W_{j}$};
		
		\node at (-3,3) [below=5mm, right=-2.2mm,name=sigma1] {\large $\sigma$};
		\node at (-3,3)[below=3mm,right=-1.9mm]{\textbullet};
		\node at (-7,5) [below=0mm, right=-3mm,name=sigmax1] {\large $\sigma_{X}$};
		\node at (1,5) [below=0mm, right=-4mm,name=sigmaw1] {\large $\sigma_W$};
		\node at (-7,5)[below=0.2mm, right=3mm]{\textbullet};
		\node at (1,5)[below=0.2mm, right=-7mm]{\textbullet};
		\draw[->] (sigmax1) -- node {}(X1);	
		\draw[->] (sigmaw1) -- node {}(W1);	
		\draw[->] (X1) -- node {}(Y1);
		\draw[->] (W1) -- node {}(Y1);
		\draw[->] (sigma1) -- node {}(Y1);
		\draw [color=gray,thick](-6,3) rectangle (-2,7);
		\draw [color=gray,thick](-4,3.5) rectangle (0,7.5);
		\node at (-6,3) [above=5mm, right=0mm] {$i \in [0,n]$};\
		\node at (0,3.5) [above=5mm, right=-15mm] {$j \in [0,p]$};
		
		\node at (-7,8.3) [below=0mm, right=-4mm] {\Large (a)};
		\end{tikzpicture}
		\caption{Illustration of (a) PMF and (b) LGMD.}
		\label{fig:ill_lgmd}
\end{figure*}
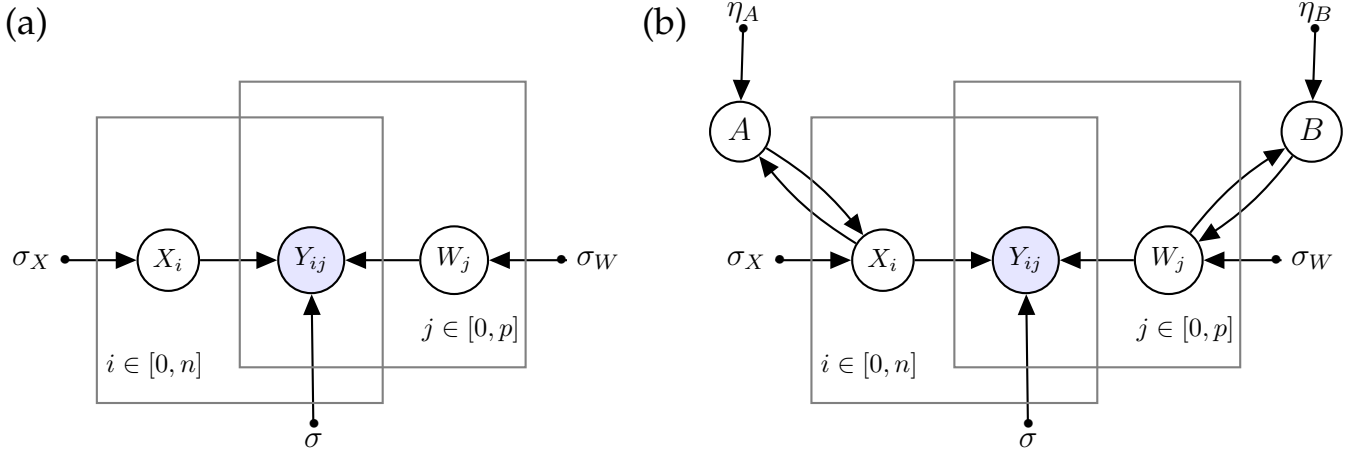

\par To estimate a large precision matrix, the graphical Lasso model has been developed by employing the sparsity assumption, i.e., many entries in the precision matrix are zeros. Specifically, it adds the $l_1$-norm penalty onto the precision matrix $\Theta$ as follows
\begin{equation}
		\ln |\Theta| -\tr(S \Theta) -\rho \norm{\Theta}_1
\end{equation}
where $\rho > 0$ controls the sparsity of $\Theta$. This problem has been extensively studied and many algorithms have been proposed \cite{yuan2007model, friedman2008sparse, li2010inexact, hsieh2014quic, danaher2014joint, cai2016estimating}. Readers may refer to \cite{fan2016overview} for a comprehensive review. However, these graphical Lasso algorithms are computationally expensive for large-scale problems. To address this, a fast heuristic method based on simply thresholding the sample covariance matrix has been developed \cite{sojoudi2016equivalence,fattahi2017graphical}. Furthermore, it has been proved that there is an explicit closed-form solution under some less stringent conditions when the graphs are sufficiently sparse.

\section{LGMD}
We propose the LGMD model to simultaneously consider the structures of both the sample and feature manifolds (Fig. \ref{fig:ill_lgmd}b). Furthermore,  we develop a block-coordinate optimization scheme based on the iterative updating rules of two low-rank factor matrices and the estimation of their corresponding precision matrices to solve it.
\subsection{Model Construction} \label{headings}
Let's first recall the generative model of PPCA (Fig. \ref{fig:ill_lgmd}a)
\begin{equation}
		Y = XW^T + E
\end{equation}
The underlying noise matrix $E$ is assumed to be IID Gaussian. Thus, the conditional distribution over the observations can be defined as
\begin{equation}
	p(Y|X,W,\sigma^{2})=\prod_{i=1}^{n}\prod_{j=1}^{p}\mathcal{N}(Y_{ij}|X_{i}W_{j}^{T},\sigma^{2}I)
\end{equation}
where $\mathcal{N}(x|\mu,\sigma^{2})$ is the probability density function of the Gaussian distribution with mean $\mu$ and variance $\sigma^{2}$. In order to capture the distribution of the latent variables, a naive approach is to simply assume that the latent variables follow a multivariate Gaussian distribution, i.e., $\text{vec}(X) \sim \mathcal{N}(\text{vec}(\textbf{X}),\Sigma_X),\text{vec}(W) \sim \mathcal{N}(\text{vec}(\textbf{W}),\Sigma_W)$. However, $X$ and $W$ consist of $k(n+p)$ variables in total, the corresponding covariance matrices are of size $nk \times nk$ and $pk \times pk$ respectively, which are too huge to estimate.

\par A more reasonable assumption is that the latent variables are correlated in the sample and feature spaces, respectively. Thus, we place the matrix Gaussian priors over $X$ and $W$ as follows
\begin{equation}
	\left\{\begin{array}{l}{p(X|A)=\mathcal{MN}_{n,k}(X|0, A^{-1},\sigma_{X}^{2}I)} \\ {p(W^{T}|B)=\mathcal{MN}_{k,p}(X|0,\sigma_{W}^{2}I,B^{-1})}\end{array}\right.
\end{equation}
where $A_{n\times n}$ and $B_{p\times p}$ are among-sample and among-feature inverse covariance matrices, i.e., precision matrices, respectively. Additionally, the columns of $X$ and $W$ are assumed independent and identically distributed, and the variance are $\sigma_{X}^{2}$ and $\sigma_{W}^{2}$ respectively. Note that the matrix normal is the same as multivariate normal distribution if and only if
\begin{equation}
	\left\{\begin{array}{l}{\text{vec}(X)\sim \mathcal{N}_{nk}(0,A^{-1}\otimes \sigma_{X}^{2}I)} \\ {\text{vec}(W^{T})\sim \mathcal{N}_{kp}(0,\sigma_{W}^{2}I\otimes B^{-1})}\end{array}\right.
\end{equation}
where $\otimes$ denotes the Kronecker product. Therefore, the log of the posterior distribution over $X$ and $W$ is given by
\begin{align*}
	\begin{split}
		&\ln p(X,W|Y,A,B,\sigma^{2})=-\frac{1}{2\sigma^{2}}\sum_{i=1}^{n}\sum_{j=1}^{p}(Y_{ij}-X_{i}W_{j}^{T})^{2}\\
        &-\frac{1}{2\sigma_{X}^{2}}\mbox{tr}(X^{T}AX)+\frac{k}{2}\ln|A|-\frac{1}{2\sigma_{W}^{2}}\mbox{tr}(W^{T}BW)\\
        &+\frac{k}{2}\ln|B|+C
	\end{split}
\end{align*}
where $|M|$ is the determinant of a matrix $M$, and $C$ is a constant that does not depend on the parameters. Maximizing the log-posterior with fixed hyperparameters (noise variance and prior variances) is equivalent to minimizing the sum-of-squared-errors function with quadratic regularization terms
\begin{align}
	\begin{split}
		\label{eq:likelihood}
		\mathcal{L} =
        & \frac{1}{2}||Y-XW^{T}||_{F}^{2}+\frac{\lambda_1}{2}(||A^{\frac{1}{2}}\hat{X}||_{F}^{2}-k\ln|A|)\\
        &+\frac{\lambda_2}{2}(||\hat{W}^{T}B^{\frac{1}{2}}||_{F}^{2}-k\ln|B|)
	\end{split}
\end{align}
where $\lambda_{1}=\sigma^2/\sigma^2_X$ and $\lambda_{2}=\sigma^2/\sigma^2_W$. $\hat{X}$ and $\hat{W}$ are normalized $X$ and $W$ by dividing their standard deviations respectively, which will be denoted as $X$ and $W$ for simplicity in the rest of this paper. Under the matrix normal distribution assumption, we reduce the number of parameters of the covariance from $(n^2+p^2)k^2$ to $n^2 + p^2$. However,  it is still infeasible to  minimize the negative log-likelihood in Eq. (\ref{eq:likelihood}). Because the number of free parameters in $A$ and $B$ grows quadratically with $n$ and $p$, respectively. More importantly, the underlying graphical structure of sample and feature representations should be very sparse. Thus, we impose sparse constraints onto the precision matrices $A$ and $B$ by introducing $l_1$-norm regularizations
\begin{align}
	\begin{split}\label{eq:objective}
		\mathcal{L} =& \frac{1}{2}||Y-XW^{T}||_{F}^{2} \\
         &+\frac{\lambda_1}{2}(||A^{\frac{1}{2}}X||_{F}^{2}-k\ln|A|+\eta_{1}^{'}\|A\|_{1,\text{off}})\\
         &+\frac{\lambda_2}{2}(||W^{T}B^{\frac{1}{2}}||_{F}^{2}-k\ln|B|+\eta_2^{'}\|B\|_{1,\text{off}})
	\end{split}
\end{align}
where $\norm{\cdot}_{1, \text{off}}$ represents the off-diagonal $l_1$-norm, $\eta_1^{'}$ and $\eta_2^{'}$ control the sparsity of $A$ and $B$, respectively. When $\eta_{1}^{'}$ and $\eta_{2}^{'}$ are too large, $A$ and $B$ turns to be diagonal. Thus, PMF can be viewed as a special case of LGMD.

\begin{figure}
	\centering
	\includegraphics[width=8.0cm]{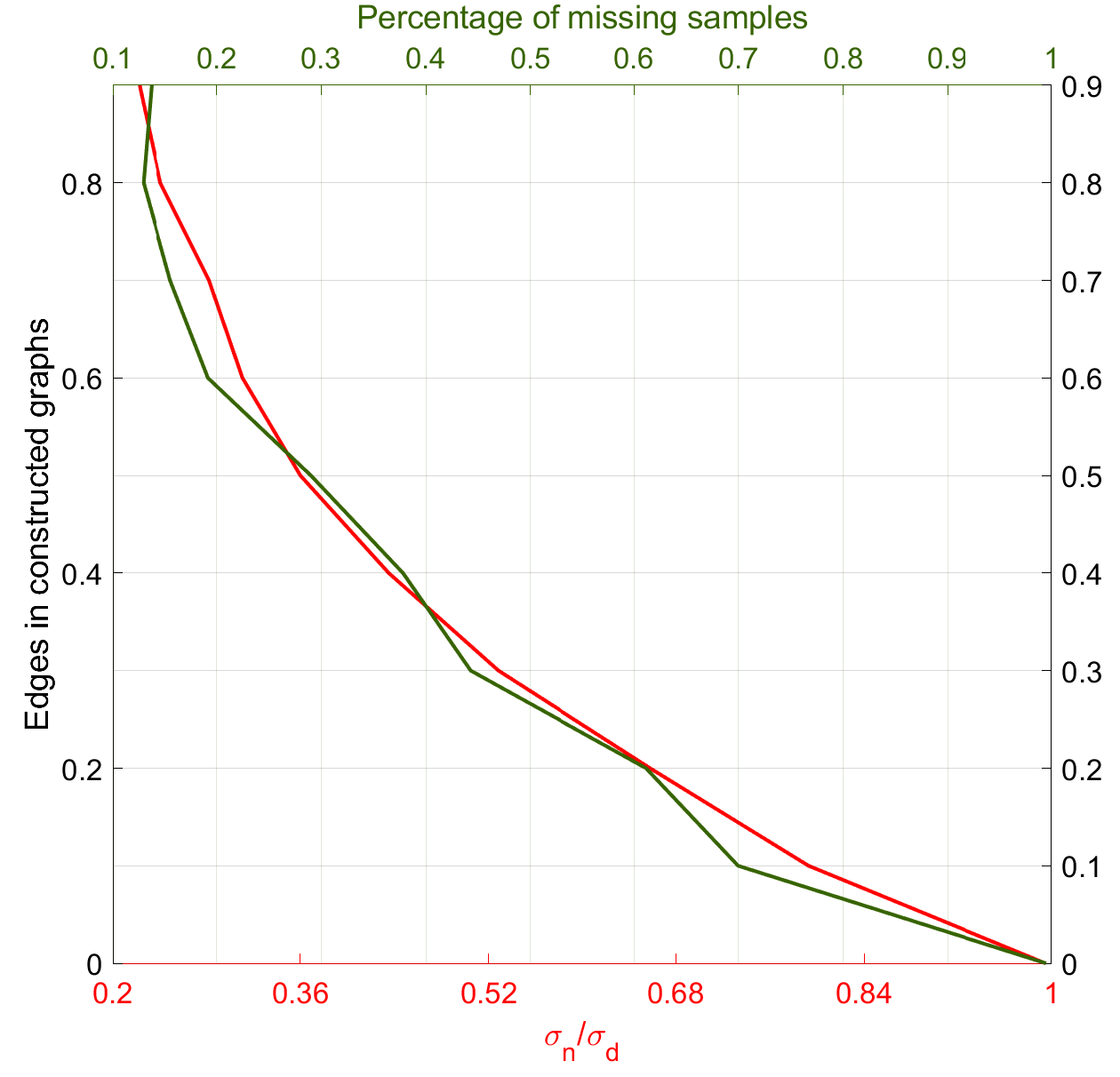}
	\caption{Change of the graph Laplacian with the increase of the noise level and the percentage of missing entries. The blue and red lines represent the change of edge numbers under different noise levels $\sigma_n$ with respect to (w.r.t.) the data standard deviation (STD) $\sigma_d$ and missing entries percentages, respectively.}
	\label{fig:noisy_laplacian}
\end{figure}

\subsection{Model Interpretation}
A common assumption in GRMD is that the latent variables $X$ and $W$ have the same structures as the original matrix data $Y$. Dual GRMD (dGRMD) considers the following formula
\begin{equation}
	\norm{Y-XW^T}_{F}^{2}+\lambda_{1}\tr(X^TLX)+\lambda_2\tr(W^TL_cW)
\end{equation}
where $L$ and $L_c$ are the graph Laplacian of sample space and feature space respectively, and they both are computed through the original data matrix $Y$ or the prior graph structure. However, this is often not easy when the data is relatively noisy and incomplete. As we can see in Eq. (\ref{eq:objective}), when $A$ and $B$ are assumed to be known, LGMD becomes the conventional dGRMD with the graphs being the precision matrices of $X$ and $W$ respectively.

\par Another application of LGMD is matrix completion, which is the task of filling in the missing entries of a partially observed matrix. One classical example is the movie-rating matrix such as the Netflix problem. Given a rating matrix in which each entry $(i,j)$ represents the rating of movie $j$ by customer $i$ if customer $i$ has rated movie $j$ and is otherwise missing, we want to predict the remaining entries in order to make good recommendations to customers on what to watch next. A common method for matrix completion is based on GRMD as follows
\begin{equation*}
	\sum_{i,j}{O_{ij}(Y_{ij}-X_{i}W_{j}^T)}^{2}+\lambda_{1}\tr(X^TLX)+\lambda_2\tr(W^TL_cW)
\end{equation*}
where $O_{ij}$ is an indicator matrix of $Y$, which means $O_{ij} = 1$ if $Y_{ij}$ is observed and 0 otherwise. The construction of the graph based on the partially observed data is a problem. Both data noise and missing entries have great impact on its estimation (Fig. \ref{fig:noisy_laplacian}).

\begin{figure}
	\includegraphics[width=8.0cm]{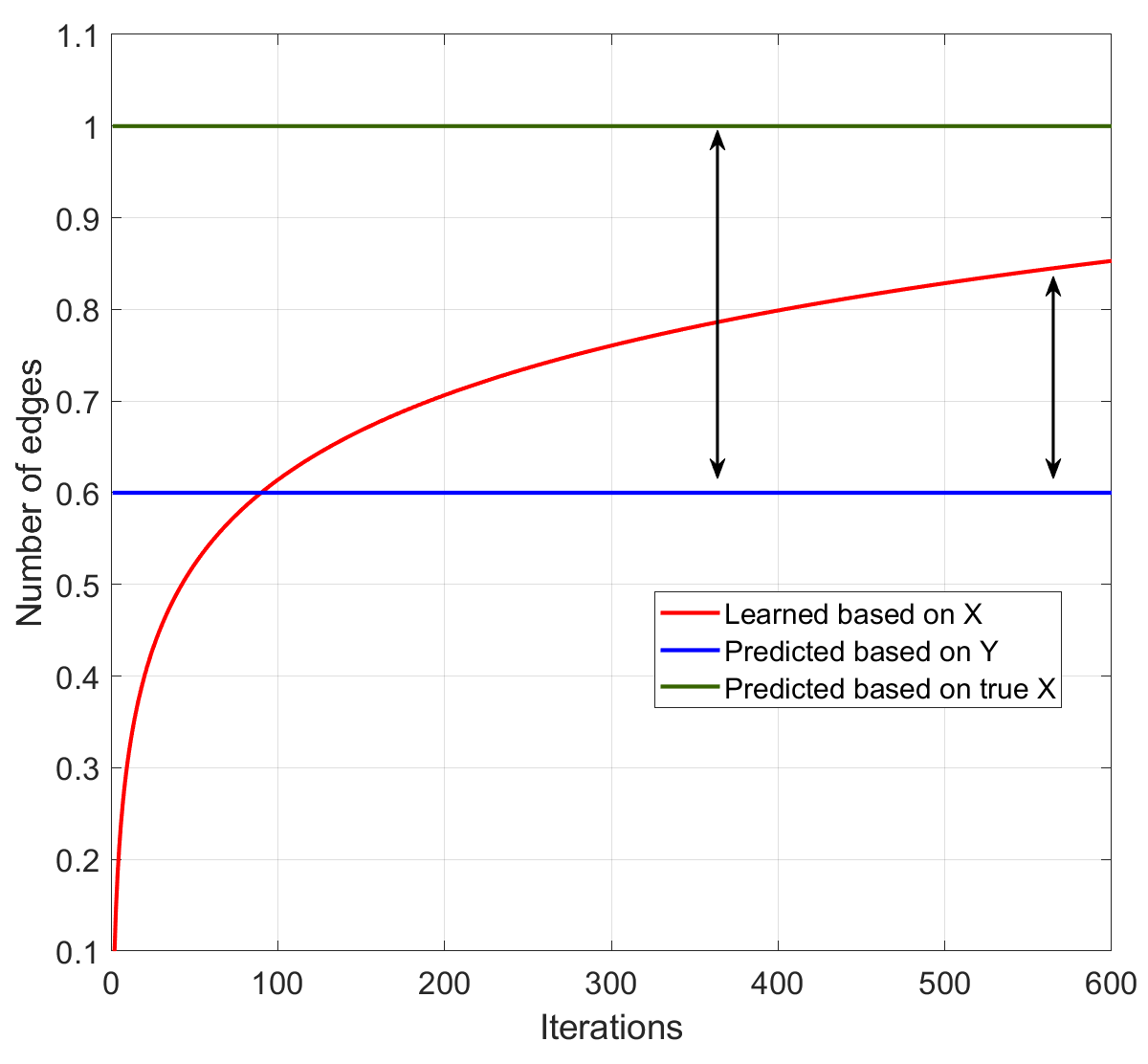}
	\caption{The number of learned edges based on the latent variables $X$ versus the iterations.}
	\label{fig:structs}
\end{figure}

In order to reduce the influence of the bias of the graph estimation, LGMD attempts to learn and modify the graphs in the iterations. We know that even if without the regularization terms, minimizing the objective function $ \min_{X,W} \norm{Y-XW^T}_{F}^{2}$ can make $X$ and $W$ automatically learn the structure of the original data to some extent (Fig. \ref{fig:structs}), which can be used to modify the initial estimated graphs and accelerate the decomposition.

\subsection{Maximum Regularized Likelihood}
Clearly, minimizing the objective function in Eq. (\ref{eq:objective}) is a matrix optimization problem involving four matrix variables $X$, $W$, $A$ and $B$. $X$ and $W$ are controlled by the graphs $A$ and $B$, while $A$ and $B$ are estimated based on $X$ and $W$ under the sparsity assumption. It is easy to verify that the function Eq. (\ref{eq:objective}) is convex relative to each matrix variable, respectively. Thus, a local minimum can be found by the block coordinate descent \cite{06222e97778647fe8c84b1a04c65cde4, Cui2019L21GRMFAI} as stated below.
\\
\textbf{Updating $X$ and $W$}: Given $A$ and $B$, one retains the terms involving $X$ and $W$
\begin{equation}
	\label{eq:objective_XW}
	\mathcal{L}_{1} = \frac{1}{2}||Y-XW^{T}||_{F}^{2}+\frac{\lambda_1}{2}||X^TAX||_{F}^{2}+\frac{\lambda_2}{2}||W^{T}BW||_{F}^{2}
\end{equation}
LGMD becomes a dGRMD problem with the graphs $A$ and $B$ being the precision matrices of $X$ and $W$. ALS is adopted to obtain the optimal $X$ and $W$ for a local minimum
\begin{equation}
	\begin{aligned} X &= \left(YW-\lambda_{1}AX\right)\left(W^{T}W+\epsilon I\right)^{-1} \\ W^{T} &=\left(X^{T}X+ \epsilon I\right)^{-1}\left(X^{T}Y-\lambda_2W^TB\right) \end{aligned}
\end{equation}
where $\epsilon$ is a small positive constant to protect the inverse of matrices from singular. $\left(X^{T}X+ \epsilon I\right)^{-1}$ and $\left(W^{T}W+\epsilon I\right)^{-1}$ are pretty small matrices and easy to compute.\\
\textbf{Updating $A$ and $B$}: A straightforward approach is to optimize $A$ and $B$ iteratively as follows
\begin{equation}
	\label{eq:precision}
	\left\{\begin{array}{l}{\hat{A}=\arg \min _{A} \operatorname{tr}\left(A S_{1}\right)-\log |A|+\eta_{1}\left\|A\right\|_{1}} \\ {\hat{B}=\arg \min _{B} \operatorname{tr}\left(B S_{2}\right)-\log |B|+\eta_{2}\left\|B\right\|_{1}}\end{array}\right.
\end{equation}
where $\eta_1 = {\eta_1^{'}}/{k}$ and $\eta_2 = {\eta_2^{'}}/{k}$, $S_{1}={XX^{T}}/{k}$ and $S_{2}={WW^{T}}/{k}$ are sample and feature space covariance matrices, respectively. Efficient optimization algorithms have been intensively studied as we described in Section \ref{sec:sparse_precision} for such $l_{1}$-norm regularized precision matrix estimation problems. We use an explicit closed-form solution based on simply thresholding the covariance matrices, denoted as \texttt{Threshold}, as a basic solver \cite{8431654,pmlr-v80-zhang18c}. Take $A$ as an example. The approximate closed-form solution for the graphical Lasso is as follows
\begin{small}
	\begin{equation}
	\label{eq:thresh}
	A_{i j}=\left\{\begin{array}{ll}{\frac{1}{\Sigma_{i i}}\left(1+\sum\limits_{(i, m) \in \mathcal{E}^{\mathrm{res}}}^{ } \frac{\left(\Sigma_{im}^{\mathrm{res}}\right)^{2}}{\Sigma_{ii}\Sigma_{mm}-(\Sigma_{im}^{\mathrm{res}})^{2}}\right)} & {\text {if } i=j} \\
{\frac{-\Sigma_{i j}^{\mathrm{res}}}{\Sigma_{i i} \Sigma_{j j}-\left(\Sigma_{i j}^{\mathrm{res}}\right)^{2}}} & {\text {if } (i, j) \in \mathcal{E}^{\mathrm{res}}} \\ {0} & {\text {otherwise}}\end{array}\right.\\
	\end{equation}
\end{small}
where $\Sigma$ is the empirical covariance matrix and $\Sigma^{\mathrm{res}}$ is short for $\Sigma^\mathrm{res}(\lambda)$, which is defined as the residual of $\Sigma$ relative to $\lambda$. The $(i,j)$-th entry of $\Sigma^\mathrm{res}(\lambda)$ equals to $\Sigma_{i j}-\lambda \times \mbox{sign}(\Sigma_{ij})$ if $i \neq j \text { and }\left|\Sigma_{i j}\right|>\lambda$, and 0 otherwise. $\mathcal{E}^{\mathrm{res}}$ is the support graph of $\Sigma^\mathrm{res}$, which indicates the edge of $(i,j) \in \mathcal{E}^{\mathrm{res}}$ if $\Sigma_{i j}^{\mathrm{res}} \neq 0$. This formula can be applied to solve $A$ and $B$
\begin{equation}
	\left\{
	\begin{aligned} \label{eq:flip_flop_l1}
			A = \texttt{Threshold}(S_1, \eta_1) \\
			B = \texttt{Threshold}(S_2, \eta_2)
	\end{aligned}
	\right.
\end{equation}
The inference scheme is summarized in \textbf{Algorithm} \ref{algo:lgmd}.\\
\textbf{Post processing:} Given any invertible matrix $S$, $XSS^{-1}W^T$ equals $XW^T$. Therefore, $X$ and $W$ are not identifiable. To obtain a unique low-rank representation, we use GPCA \cite{allen2014generalized}  to post-process $X$ and $W$. Specifically, given $A^{-1}$ and $B^{-1}$, GPCA solves the following problem
\begin{align}
		\begin{split}\label{eq:gpca}
			\min & \frac{1}{2} \norm{Y - UDV^T}_{A^{-1}, B^{-1}}^2\\
			\text{s.t. }    & U^TB^{-1}U = I\\
			& V^TA^{-1}V = I \\
			& \text{diag}(D) \geq 0
		\end{split}
\end{align}
where $\norm{X}_{A^{-1}, B^{-1}} = \sqrt{A^{-1}XB^{-1}X^T}$.
It has been proved that GPCA converges to the unique global solution when $A^{-1}$ and $B^{-1}$ are positive-definite. Then we set $X=UD$ and $W=V$ as the output.\\

Additionally, as mentioned in the last part, a positive off-diagonal entry in the precision matrix implies a negative partial correlation between the two random variables, which is difficult to interpret in some contexts, such as road traffic networks \cite{7979524,Dong2019LearningGF,Pavez2016GeneralizedLP}. For such application settings, it is therefore desirable to learn a graph with non-negative weights. The problem has been formulated as \cite{Lake10discoveringstructure}
\begin{align*}
	\label{eq:gsp}
	& \mathop{\text{max}}\limits_{\Theta,\sigma} \ln |\Theta| -\tr(S \Theta) -\rho \norm{\Theta}_1 \\
	&\text{s.t.  } \Theta = L + \frac{1}{\sigma^2}I, L \in \mathcal{L}
\end{align*}
where $I$ is the identity matrix, $\sigma^2$ is the a priori feature variance, $\mathcal{L}$ is the set of valid graph Laplacian matrices. The solution was found using the package CVX for solving convex programs \cite{cvx}. We denote the proposed algorithm estimating graphs with the above formula as LGMD$+$ and we will apply it to the real-world scenarios for comparison.
\begin{table}
	\begin{minipage}{\columnwidth}
		\begin{algorithm}[H]
			\begin{algorithmic}[1]\caption{\textbf{Maximum Regularized Likelihood}}\label{algo:lgmd}
					\Input data matrix $Y$, rank $k$
					\Output  $X$, $W$, $A$ and $B$
					\State Truncated $k$ SVD of $Y=U_k \Sigma_k V_k$
					\State Initialize $X=U_k$ and $Y=\Sigma_k V_k$, $A = I$, $B=I$
					\Repeat
					\State update $A$ and normalize $A$
					\State update $B$ and normalize $B$
					\Repeat
					\State update $X$
					\State update $W$
					\Until{convergence}
					\Until{Change of the objective function value is small enough}
			\end{algorithmic}
		\end{algorithm}
	\end{minipage}
\end{table}

\subsection{Computational Complexity}\label{sec:complexity}
\par Here we briefly discuss the computational complexity of the maximum regularized likelihood algorithm for LGMD. At each iteration of updating $X$, the computational cost lying in matrix multiplication and inverse is $O(n^2k + npk)$. Similarly, the computational cost for updating $W$ is $O(p^2k + npk)$.

\par Generally, solving $A$ and $B$ is more computationally expensive if we use conventional algorithms. Thus, we adopt the approximate thresholding method, which gives a close enough solution and is less time-consuming than traditional methods. Given two empirical covariance matrices of $n \times n$ and $p \times p$, the computational complexities of solving $A$ and $B$ are $O(n^3)$ and $O(p^3)$, respectively. In contrast, solving the constrained precision estimation is more time-consuming, which lacks an effective thresolding algorithm. We will compare the running time in the real-world data experiment.

\begin{figure*}[!t]
    \centering
    \includegraphics[width=0.9\textwidth]{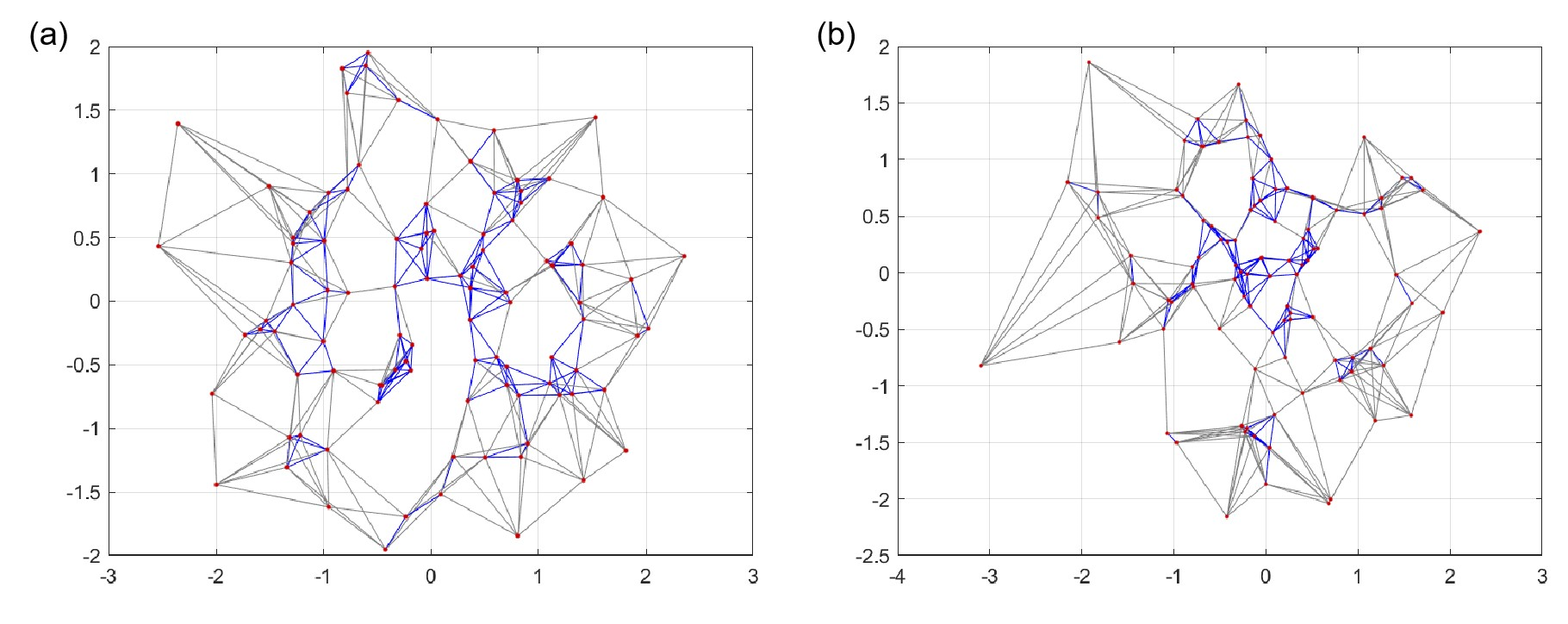}
	\caption{The support graphs of the two generated ground-truth precision matrices (a) $A$ and (b) $B$ using a data with 100 samples and 100 features with dimension reduction. The blue and the gray lines represent positive and negative edges, respectively.}
	\label{fig:syn_xwstruct}
\end{figure*}
	
\subsection{Tuning Parameter Selection}
\par In our algorithm, we have four hyperparameters in total, in which $\lambda_1$ and $\lambda_2$ balance the trade-off of the low-rank approximation and structural restoration while $\eta_1$ and $\eta_2$ control how many edges should we restore. In the experiments, we kept the sparseness of the graphs fixed, since one of our goals is to test whether the learned graphs are informative. We only tune two parameters $\lambda_1$ and $\lambda_2$. However, it still takes a long time (minutes or hours) to evaluate using the grid-search strategy. Thus, we adopt an effective Bayesian optimization approach to optimize, which uses fewer combinations of hyperparameters than the grid-search strategy. Bayesian optimization is best-suited for optimization over continuous domains of less than 20 dimensions, and tolerates stochastic noise in function evaluations \cite{frazier2018tutorial}. We also applied the Bayesian optimization strategy to other methods in the experiments.

\section{Experimental Results}  \label{experiments}
In this section, we demonstrate the effectiveness of LGMD on both synthetic and real-world data and compare it with PCA, PMF and dGRMD. With the synthetic datasets, we demonstrate the capability of LGMD to improve denoising performance, and successfully infer the underlying structure. In addition, we evaluate the ability of LGMD and LGMD+ to recover the true underlying signals from noisy and incomplete samples on two sensor applications including the temperature data and netflix data. At last, we apply LGMD to real labeled image and text datasets to check its clustering performance.

\subsection{Synthetic Experiment}
For the synthetic experiment, the underlying graphs are known, and the recovered structure by LGMD can be quantitatively assessed. Due to the model complexity and the inherent coupling, these matrices can not be drawn independently, and need to be designed carefully.

\par We generated two graphs with 100 nodes each and keeping about 6$\%$ of the overall edges (Fig. \ref{fig:syn_xwstruct}), whose adjacency matrices are denoted as $A_{gt}$ and $B_{gt}$, which are symmetric and sparse as assumed. The toy data can be constructed in the following way
	\begin{align*}
		\begin{split}
		    &X_{gt}\sim \mathcal{MN}_{n,k}(0,A^{-1}_{gt},I) \\
            &W_{gt}^{T}\sim \mathcal{MN}_{k,p}(0,I,B^{-1}_{gt}) \\
		    &Y_{gt} = X_{gt}W_{gt}^{T} \\
			&Y_{no} = Y_{gt} + E
		\end{split}
	\end{align*}
where the entries of $E$ follow the Gaussian noise with variance $\sigma^{2}$, $Y_{gt}$ represents the ground truth matrix and the noise matrix is denoted as $Y_{no}$.

To test the quality of the learned representations, the final performance of each method on each experiment was measured as the average over 30 realizations with the following eight measures
\begin{align*}
		&E_1=\left\|O\odot(Y_{no}-\tilde{X}\tilde{W}^{T})\right\|_{F}, \ E_2=\left\|Y_{gt}-\tilde{X} \tilde{W}^{T}\right\|_{F},\\&E_3=\text{corr}(\tilde{X}), \ E_4=\text{corr}(\tilde{W}),\\
		&E_{5}=\text{subspace}(X_{g t},\tilde{X}), \ E_{6}=\text{subspace}(W_{g t}, \tilde{W}),\\&E_7=\text{edge}(A_{gt},\tilde{A}), \ E_8=\text{edge}(B_{gt},\tilde{B})
\end{align*}
where $\tilde{X},\tilde{W},\tilde{A},\tilde{B}$ are the outputs of LGMD. $E_1$ is used to measure the performance of matrix completion and $O$ is an indicator matrix of $Y$ which means $O_{ij} = 1$ if $Y_{ij}$ is observed and 0 otherwise. In our experiment, we used a part of the observed entries of $Y_{no}$ for training and the remaining entries (denoted by $O$) for prediction. $E_2$ is to measure the denoising performance. They are compared in terms of the root mean squared error (RMSE). Note that $E_1$ and $E_2$ are commonly used to measure the performance of representation learning. In addition, we also adopt some other meaningful measures to evaluate if the method recovers the correct subspace. subspace($X_1,X_2$) denotes the angle between two subspaces spanned by the columns of $X_1$ and $X_2$ respectively. corr($X$) is a vector denoting all the Pearson correlation coefficients between any two columns of $X$. Lastly, edge($A_1,A_2$) denotes the edge number recovered by $A_2$ in $A_1$.

\begin{figure}
   \centering
   \includegraphics[width=0.40\textwidth]{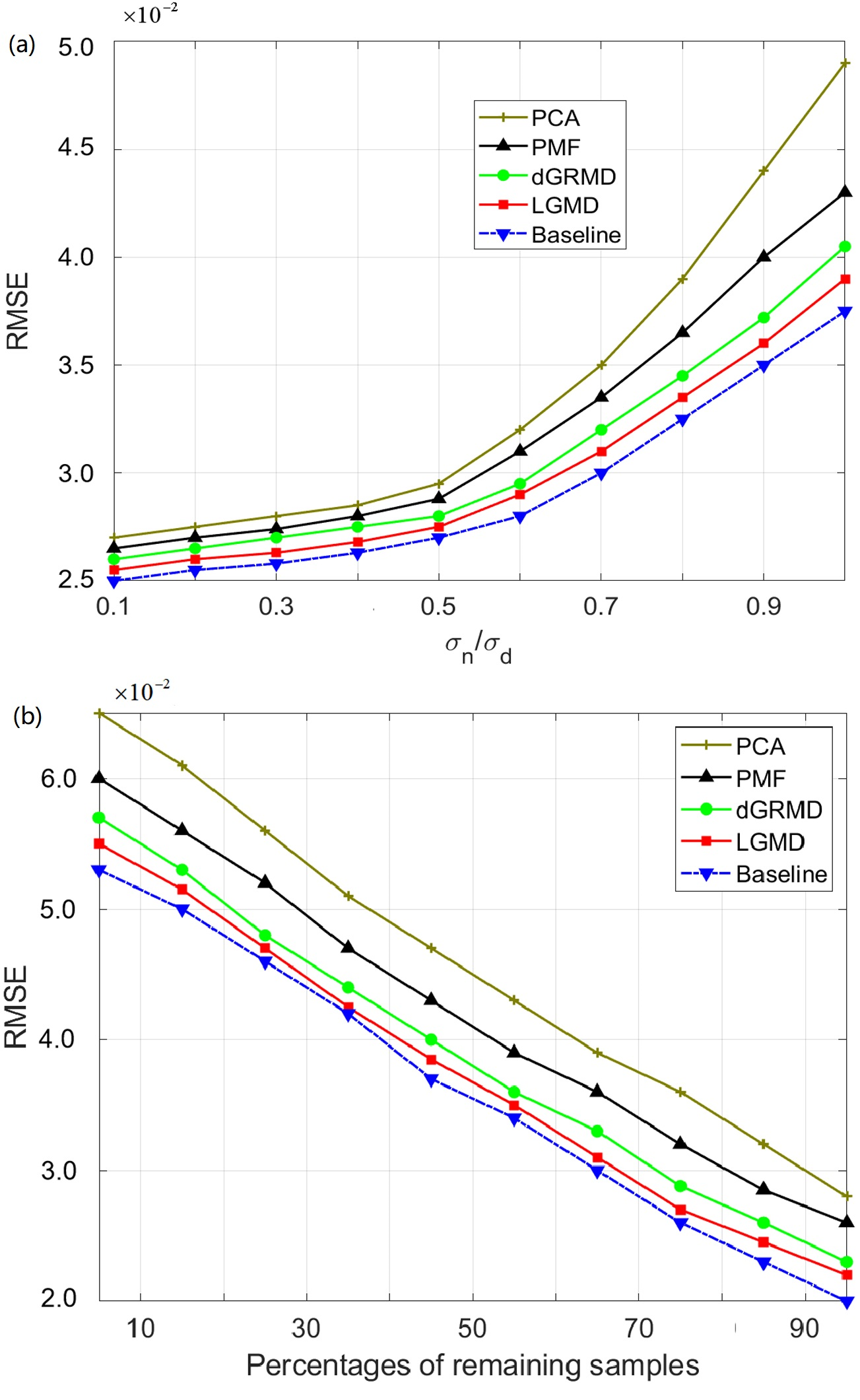}
   \caption{Comparison of four methods in terms of RMSE tested on the synthetic dataset. (a) denoising error for different noise levels $\sigma_n$ w.r.t. the data STD $\sigma_d$, and (b) data completion error for different percentages of remaining samples.}
	\label{fig:syn_e1e2}
\end{figure}

\par We will analyse the performance of the methods in the next sections with two parts. The first part mainly reports denoising and matrix completion performance ($E_1$, $E_2$), and the second part compares the ability of recovering the hidden space and structure ($E_3$-$E_8$). We use the true graphs as baseline for comparison (GRMD-real).

\begin{table*}[!t]
	\centering
	\caption{Performance of the recovered subspace angles of $X$ on the small synthetic datasets. The best results are highlighted in bold.}
	\label{table:syn_space_X}
	\begin{tabular}{lccccccccccc}
	\hline
	 $\sigma_n/\sigma_d$ & 0.1 & 0.2 & 0.3 & 0.4 & 0.5 & 0.6 & 0.7 & 0.8 & 0.9 & 1 &\\
	\hline
	PCA & 0.1004 &   0.2012  &  0.3018  &  0.4013  &  0.4977 &  0.5868  &  0.7421 &  0.8759  & 1.0333  &  1.1590 & \\
    PMF & 0.1003 &   0.2002  &  0.3014  &  0.4000  &  0.4911 &  0.5855  &  0.7398 &  0.8751  & 1.0321  &  1.1591 & \\
    dGRMD & 0.1004 & 0.2012  &  0.3016  &  0.4012  &  0.4977 &  0.5869  &  0.7408 &  0.8749  & 1.0333  &  1.1590 & \\
    LGMD  & \textbf{0.1001}  &   \textbf{0.1996}   &   \textbf{0.2974}  &   \textbf{0.3932}  &   \textbf{0.4868} &    \textbf{0.5744}  &   \textbf{0.7389} &    \textbf{0.8666}  & \textbf{1.0211}  &  \textbf{1.1551} & \\
    \hline
    GRMD-real & 0.0887 & 0.1793 & 0.2737 & 0.3725  &  0.4799 &  0.5736  &  0.6671 &  0.8049  & 0.9366  &  1.0151 & \\
    \hline
    \end{tabular}
\end{table*}
	
\begin{table*}[!t]
	\centering
	\caption{Performance of the recovered subspace angles of $W$ on the small synthetic datasets. The best results are highlighted in bold.}
	\label{table:syn_space_W}
	\begin{tabular}{lccccccccccc}
		\hline
		 $\sigma_n/\sigma_d$ & 0.1 & 0.2 & 0.3 & 0.4 & 0.5 & 0.6 & 0.7 & 0.8 & 0.9 & 1 &\\
		\hline
		PCA & 0.0883 &  0.1787 &  0.2738 &  0.3748 & 0.4802 & 0.5887 & 0.7992 & 0.9206 & 1.0557 & 1.1930 &\\
        PMF & 0.0883 &  0.1785 &  0.2741 &  0.3747 & 0.4800 & 0.5887 & 0.7991 & 0.9200 & 1.0551 & 1.1935 &\\
        dGRMD &0.0881 & 0.1782 &  0.2733 &  0.3746 & 0.4800 & 0.5885 & 0.7992 & 0.9206 & 1.0557 & 1.1930 &\\
        LGMD  & \textbf{0.0882} & \textbf{0.1762}  & \textbf{0.2676} & \textbf{0.3650} & \textbf{0.4675} &
        \textbf{0.5732} & \textbf{0.7947} & \textbf{0.9165} & \textbf{1.0441} & \textbf{1.1806} & \\
        \hline
        GRMD-real &0.0881 &0.1738 &0.2613 & 0.3412 & 0.4533 & 0.5882 & 0.7144 & 0.8728 & 0.9852 & 1.0454&\\
        \hline
   \end{tabular}
\end{table*}

\subsubsection{Reconstruction and Completion Ability}
The reconstruction error $E_1$ and data completion error $E_2$ measure the performance of the methods on denoising and missing value completion. Clearly, LGMD gets superior performance compared to three competing methods (Fig. \ref{fig:syn_e1e2}). 
dGRMD is closer to LGMD when the noise is smaller, but the results gradually get worse as the noise increases. This suggests that the learnable graph regularization in LGMD is more robust to noise and missing values.

\subsubsection{Representation Ability}
Here we explore the performance of LGMD on extracted effective representations with $E_5$ and $E_6$, which are important measures on evaluating whether the outputs truly restore the original data space (Table \ref{table:syn_space_X} and Table \ref{table:syn_space_W}). LGMD performs best in all the experiments.

\begin{figure*}[!t]
	\centering
	\includegraphics[width=0.99\textwidth]{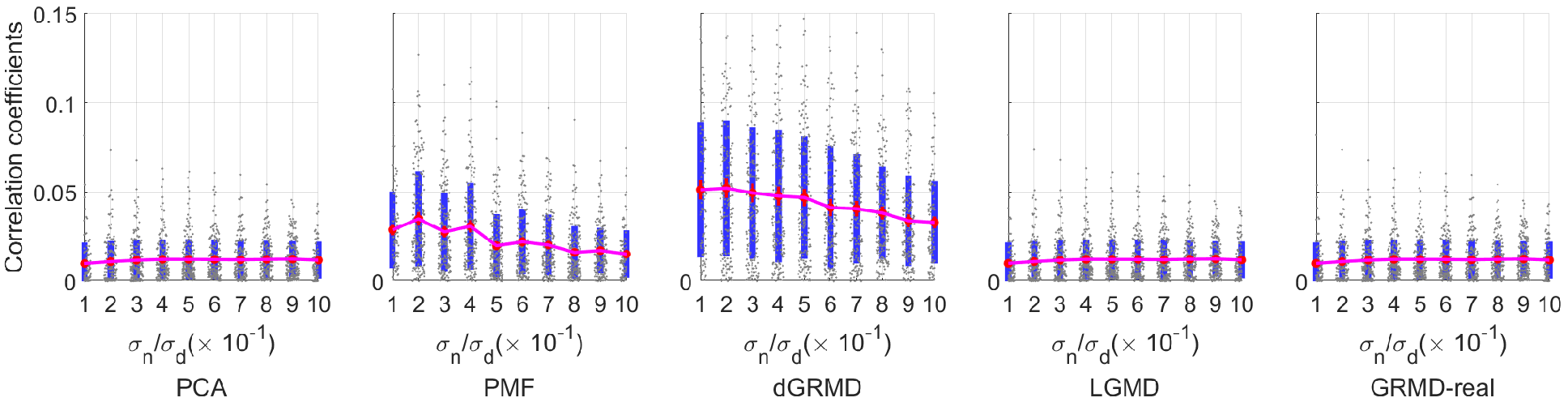}
	\caption{Comparison of four methods on the correlation coefficients. }
	\label{fig:corr}
\end{figure*}

\par \textbf{Feature decoupling} is always a hot topic and an important measure on evaluating whether the learned representation is more compact and independent \cite{8953870,6472238}. In this experiment, we generate $X$ and $W$ with rank 20, so there are 190 sets of correlation coefficients in total, which are plotted with the increase of noise (Fig. \ref{fig:corr}). The results demonstrate that LGMD has comparable performance with PCA, and will not significantly get worse as the noise increases. Whereas dGRMD and PMF fail to decouple the features.

\par
\begin{figure*}[!t]
    \centering
	\includegraphics[width=0.81\textwidth]{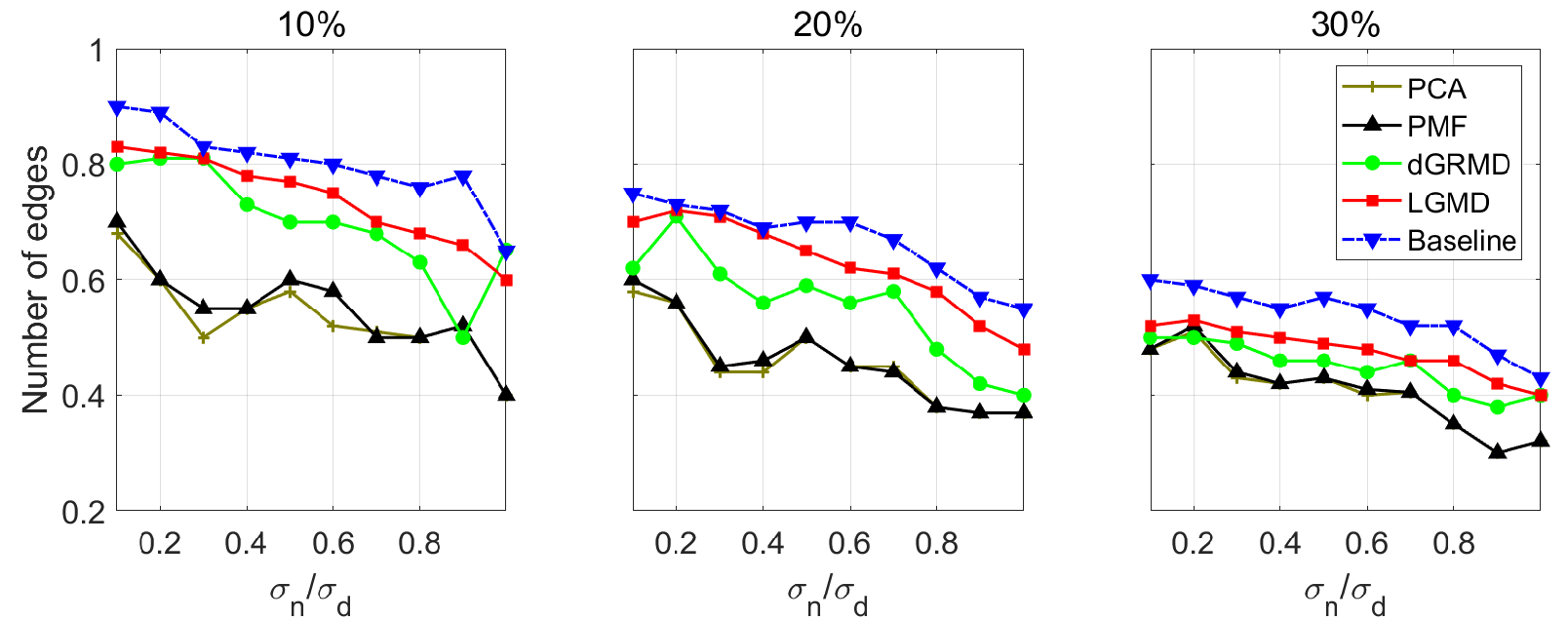}
	\caption{Comparison of four methods on the recovered structures.}
	\label{fig:syn_struct}
\end{figure*}

 A more reasonable measure is that whether the learned representation recovers the structures (edges) in the original data. So we compared the recovered structure performance (Fig. \ref{fig:syn_struct}), which illustrates the comparison results of the most significant edges from 10$\%$ to 30$\%$. The results show that our method performs best in restoring the original edges with the increase of noise level. The more significant the edges, it is more easier to be recovered by our method. Moreover, LGMD still performs best on edges that are not significant enough.

Here, we compared LGMD with several matrix decomposition methods on the measures for reconstruction, completion and representation. The results demonstrate the effectiveness of LGMD. The learnable characteristics of the graphs turns to be more robust to the noise and missing values. Furthermore, through the estimated graphs, we can get a more compact and independent representation and recover more underlying edges.

\subsection{Real-world Experiment}
This experiment aims to test the effectiveness of LGMD for data completion and denoising on real-world datasets. The underlying true graphs are unknown here. We consider two datasets including the Netflix and Temperature data, which have clear sample and feature structures. This kind of data is very common in many applications. For example, in the movie rating data, the users and movies both have their own structures. Users with similar age, gender, and growth environment tend to have similar ratings on the same movies. Then there should be a connected edge between such two users.

\par To assess the potential benefit of LGMD for denoising, Gaussian noises of different levels $\sigma_{n}^{2}$ were added to the original data. Since the noises are random and affect the graph estimation, the learnable graph-regularization is more likely to restore the true graph structure. We further evaluate the performance of LGMD for data completion. In practice, missing entries may arise either from a budget restricted data acquisition, or from faulty sensors. For this scenario, we set the noise level $\sigma_n$ = 0 and draw $O$ through randomly subsampling the test signals with various predefined percentages of samples preserved. In addition, we also demonstrate that LGMD$+$ performs best in terms of cluster accuracy, but costs more time.

\par
\begin{figure}
\includegraphics[width=0.42\textwidth]{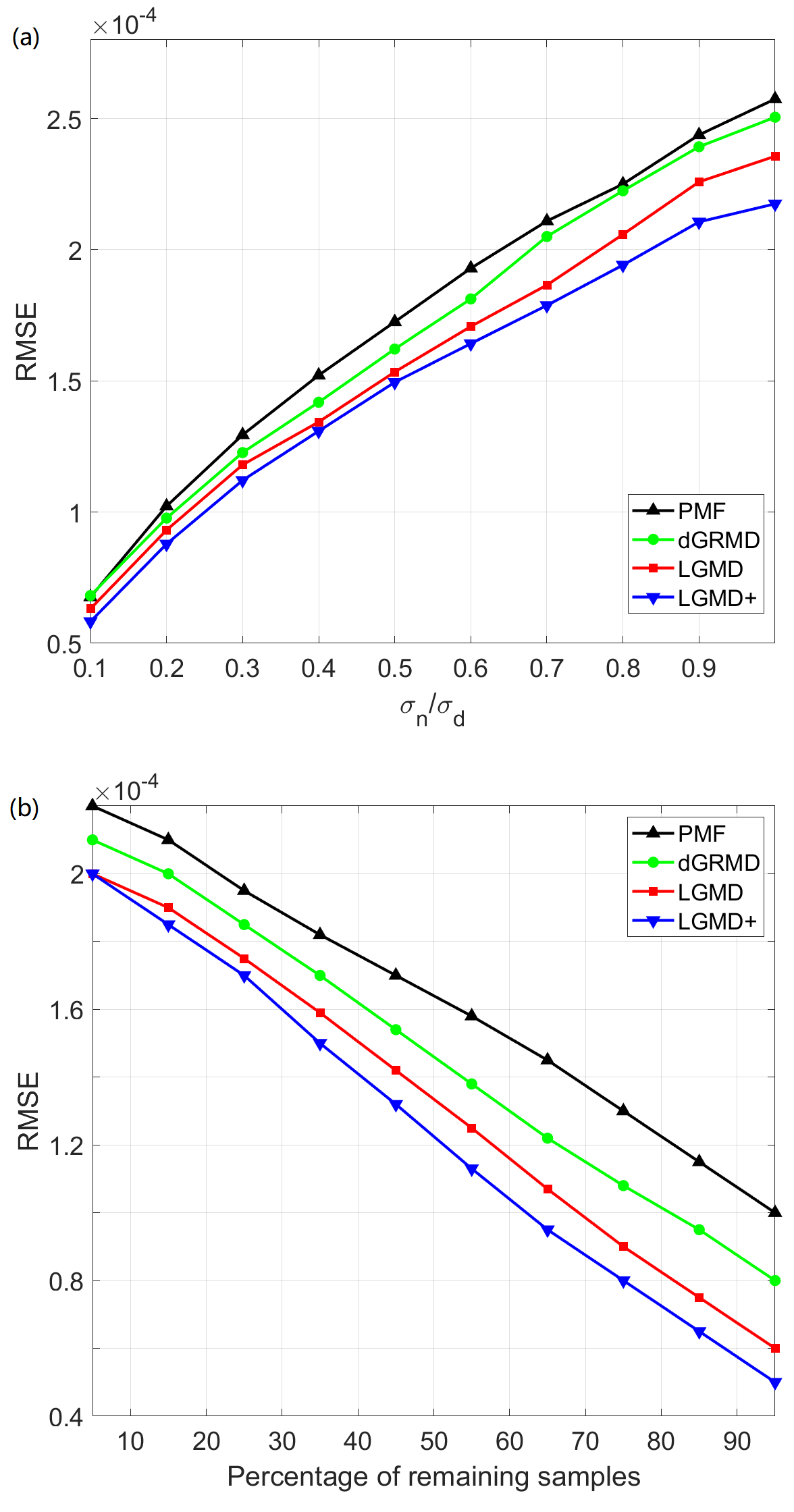}
\caption{ Comparison of four methods in terms of RMSE tested on the temperature dataset. (a) denoising error for different noise levels $\sigma_n$ w.r.t. the data STD $\sigma_d$, and (b) data completion error for different percentages of remaining samples.}    \label{Fig:temp_results}
\end{figure}

\begin{figure*}[!t]
	\centering
\includegraphics[width=0.9\textwidth]{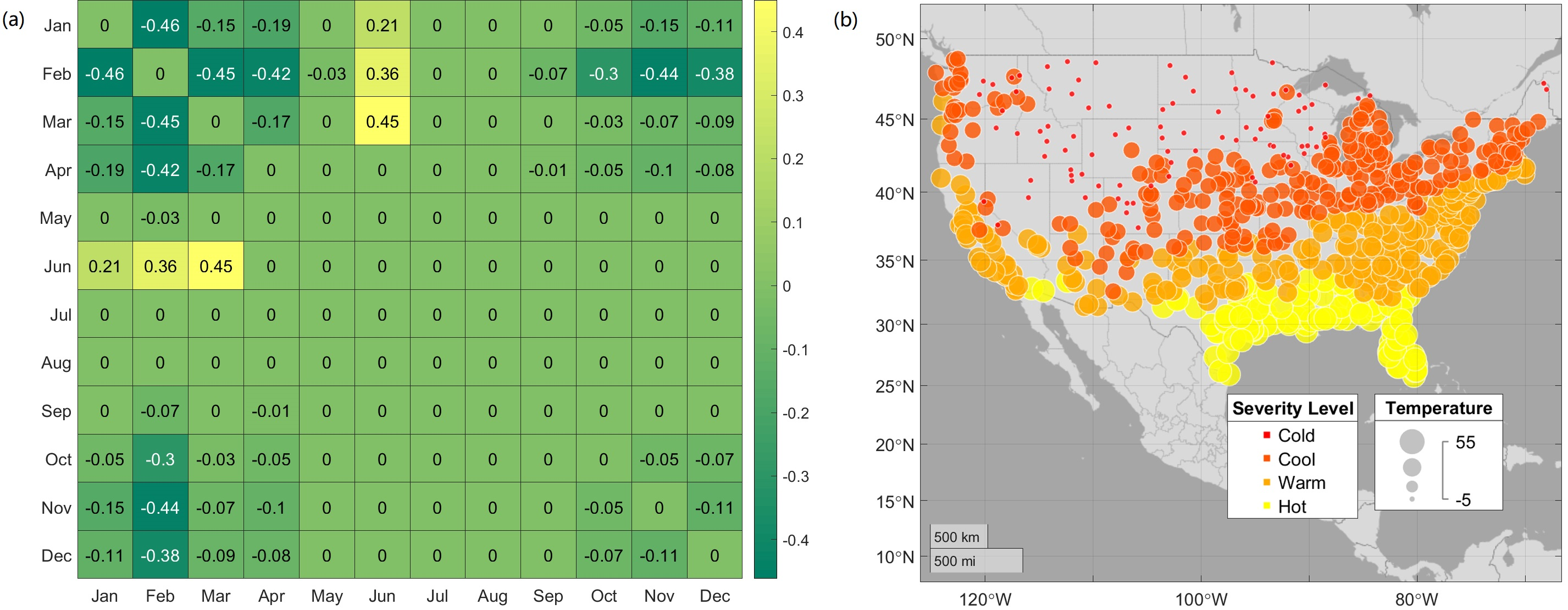}
		\caption{Illustration of the sample and feature structures of Temperature data. (a) heatmap of the temperature data of 12 months in 2018. (b) distribution of 150 sensors in the United States and the local temperature on a certain day in 2019.}
		\label{fig:temp_struct}
\end{figure*}

 \textbf{Temperature data} We first consider a dataset of daily temperature measurements collected from January 2018 to October 2019 by $N =$ 150 weather sensors across the mainland United States. The entries in the original data represents the average temperatures given in degrees of Fahrenheit measured across the sensor network on a single day. Data completion makes sense for this data because short-term temperature prediction is still problematic. The underlying assumption is that close sensors and close dates will have highly correlated temperatures. From the structure illustration (Fig. \ref{fig:temp_struct}), we can see that the temperature in spring is similar to that in winter while more different from that in summer. Moreover, from the distribution of predefined 150 sensors in the United States, sensors with similar temperatures are always distributed in similar places. Making full use of the structure of the data itself can predict future temperature. The results clearly demonstrates the advantages of LGMD and LGMD+ on temperature prediction and denoising (Fig. \ref{Fig:temp_results}). LGMD$+$ performs the best with various noise levels and missing entry percentages.

\par
\begin{figure}
        \includegraphics[width=0.42\textwidth]{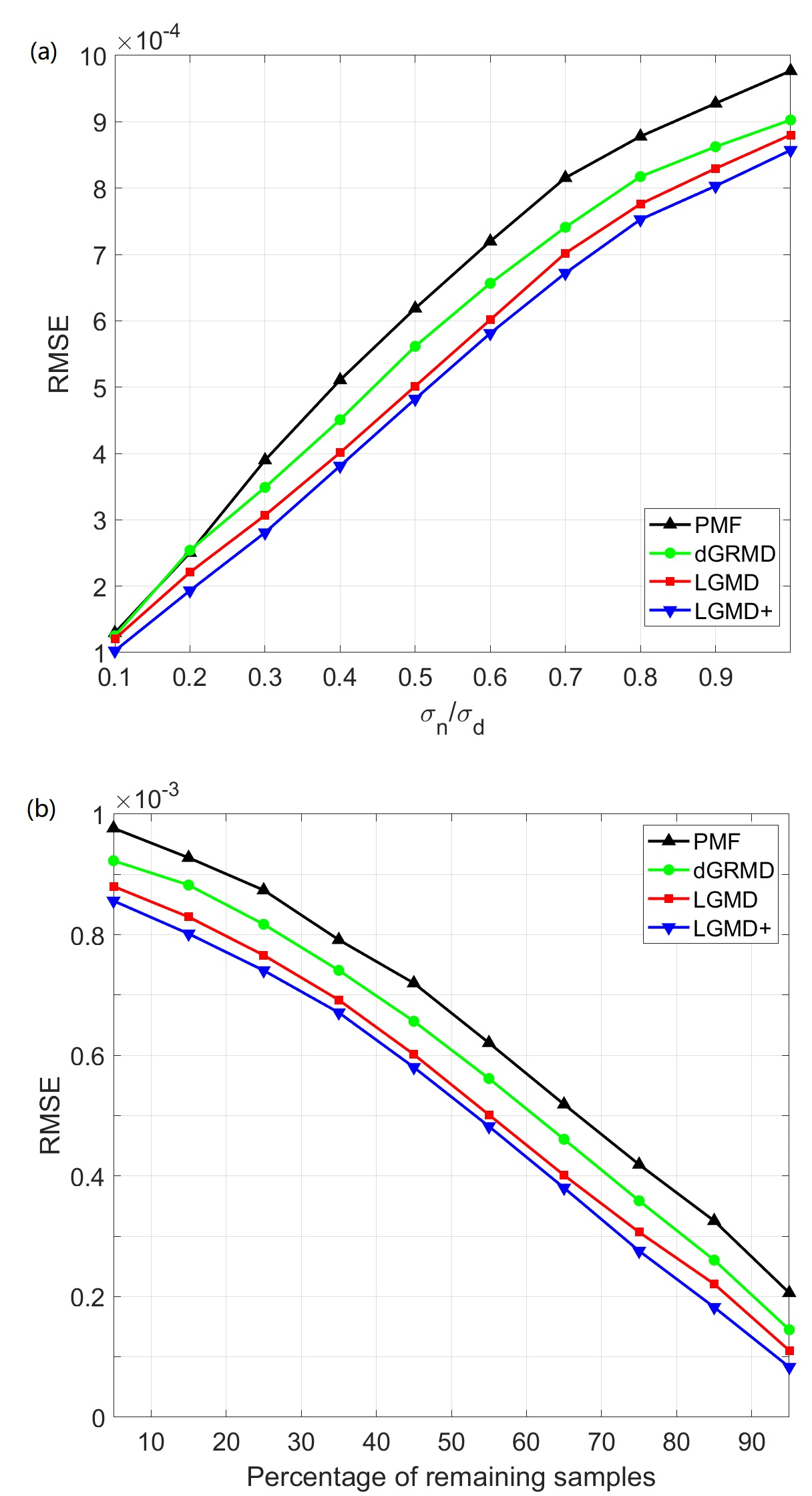}
		\caption{Comparison of four methods in terms of RMSE tested on the Netflix dataset (the same setting with Fig. \ref{Fig:temp_results}).} 
		\label{Fig:net_results}
\end{figure}

 \textbf{Netflix data} Recommendation system is a hot and difficult problem nowadays. Graph-regularization terms are often used to obtain more meaningful features, in which the graphs are constructed based on scores given by the users. However, the computation of the graphs is often biased because of the sparsity of the data. The Netflix data was used here. This data was collected between October 1998 and December 2005 and represents the distribution of all ratings. The training dataset consists of 100,480,507 ratings from 480,189 randomly-chosen, anonymous users on 17,770 movie titles. 1000 users and movies were chosen respectively to conduct this experiment for convenience. The numerical results here further demonstrated the superiority of LGMD and LGMD$+$ (Fig. \ref{Fig:net_results}). With the increase of noise level and proportion of missing entries, the performance of PMF and dGRMD gradually become worse, while LGMD and LGMD$+$ show robust performance.

\par Table \ref{table:running_time} reports the running time per 1000 iterations on denoising and completion tasks on the temperature data. The running time of LGMD is acceptable while LGMD$+$ is more time-consuming compared to other methods. 

\begin{table}[]
\centering
\caption{Comparison of Running Time on the Temperature data}
\begin{tabular}{lllll}
\hline
    & PMF & dGRMD & LGMD & LGMD$+$ \\ \hline
Denoising & 10s & 20s   & 230s  & 2500s\\ \hline
Completion & 80s & 80s   & 500s  & 3000s \\ \hline
\label{table:running_time}
\end{tabular}
\end{table}

\par In short, LGMD not only performs better in denoising, but also outperforms PMF and dGRMD in data completion. Even with relatively high noise levels and severe missing percentages, the performance of LGMD is not seriously degraded compared to other methods, suggesting its robustness against missing data and noise. LGMD and LGMD$+$ perform the best than all the other methods, and LGMD$+$ is more time-consuming.

\subsection{Clustering Experiment}
Clustering performance is a critical measure on evaluating the learned representations. To demonstrate how the clustering performance can be improved by LGMD, we compare the following 3 popular matrix decomposition methods including $k$-means, PMF and dGRMD. Note $k$-means is indeed a special matrix decomposition method where  entries of the coefficient matrix are only 1s or 0s. Two datasets were used in this section.

\par \textbf{MNIST data} The first data set is MNIST image data consisting of 60000 gray-scale images of 32$\times$32 on hand-written digits ranging from 1 to 10.

\par \textbf{TDT2 data} The second data set is from the NIST Topic Detection and Tracking (TDT2) corpus, which consists of data collected during the first half of 1998 and taken from 6 sources, including 2 newswires (APW, NYT), 2 radio programs (VOA, PRI) and 2 television programs (CNN, ABC). Those documents appearing in two or more categories were removed and only the largest 20 categories were kept.

\par In order to facilitate the calculation, we have only used 2000 samples (Table \ref{table:mnist} and Table \ref{table:tdt2}). In order to randomize the experiments, we conduct the evaluations with different cluster numbers. For each given cluster number $K$, 20 repetitions were conducted on different randomly chosen clusters. The mean and standard error of the performance are reported. We can see that graph-regularization methods outperform PMF and $k$-means, suggesting the importance of the geometrical structure in learning the hidden factors. In addition, regardless of the data sets, LGMD and LGMD$+$ always show the best performance. This shows that by leveraging the power of learnable graph-regularization, LGMD can learn better compact representations.

\begin{table}[h!]
\centering
\setlength{\tabcolsep}{0.8mm}{
	\caption{Comparison of Clustering Accuracy on MNIST}
	\begin{tabular}{lccccc}
		\hline  
			nClusters& 10& 9& 8& 7& 6 \\
			\hline  
			$k$-means& 0.530$\pm$0.04&  0.590$\pm$0.04&    0.621$\pm$0.06&      0.657$\pm$0.04&       0.690$\pm$0.04\\
				PMF& 0.601$\pm$0.07& 0.660$\pm$0.05& 0.702$\pm$0.04& 0.740$\pm$0.04& 0.765$\pm$0.04\\
				dGRMD& 0.610$\pm$0.10& 0.677$\pm$0.05&  0.708$\pm$0.04& 0.743$\pm$0.03& 0.772$\pm$0.03\\
				LGMD& 0.640$\pm$0.08& 0.680$\pm$0.08&  0.725$\pm$0.04& 0.754$\pm$0.03& 0.790$\pm$0.02\\
				LGMD$+$& \textbf{0.648$\pm$0.05}&  \textbf{0.690$\pm$0.06}&    \textbf{0.740$\pm$0.05}&     \textbf{0.770$\pm$0.04}& \textbf{0.805$\pm$0.04}\\
			\hline
			\label{table:mnist}
	\end{tabular}}
\end{table}

\begin{table}[h!]
\centering	
\caption{Comparison of Clustering Accuracy on TDT2}
	\setlength{\tabcolsep}{0.8mm}{
	\begin{tabular}{lccccc}
	\hline  
		nClusters& 20& 18& 16& 14& 12 \\
	\hline  
				$k$-means& 0.233$\pm$0.02&  0.258$\pm$0.03&    0.266$\pm$0.04& 0.284$\pm$0.04&       0.296$\pm$0.05\\
				PMF& 0.440$\pm$0.07& 0.461$\pm$0.05& 0.505$\pm$0.04& 0.523$\pm$0.04& 0.576$\pm$0.04\\
				dGRMD& 0.447$\pm$0.04& 0.468$\pm$0.05&  0.509$\pm$0.03& 0.525$\pm$0.03& 0.598$\pm$0.03\\
				LGMD& 0.452$\pm$0.08& 0.470$\pm$0.08&  0.512$\pm$0.04& 0.528$\pm$0.07& 0.602$\pm$0.05\\
				LGMD$+$& \textbf{0.460$\pm$0.06}&  \textbf{0.478$\pm$0.06}&    \textbf{0.520$\pm$0.05}&     \textbf{0.536$\pm$0.04}& \textbf{0.615$\pm$0.05}\\
	\hline
	\label{table:tdt2}
	\end{tabular}}
\end{table}

\section{Discussion and Conclusion}

\par In this paper, we propose a novel matrix decomposition model LGMD with learnable graph-regularization to model the underlying structure in both feature and sample spaces of the latent variables by the matrix normal distribution. Intriguingly, LGMD builds a bridge between graph-regularized methods and probabilistic models of matrix decomposition. It can learn and update the graph knowledge based on the hidden information extracted from the data. To our knowledge, this is the first study to propose the concept of learnable graph-regularization and develop an effective algorithm to solve it. This method demonstrates superior performance to several competing methods on the synthetic and real-world data when the effect of the noise is mild and the percentage of missing entries is large. Specifically, numerical experiments on synthetic datasets show that considering the structures of sample and feature spaces of latent variables brings an improvement in recovering more structures, denoising and predicting missing entries. LGMD also shows its advantages in feature decoupling and subspace recovering. In addition, we also suggest to consider the non-negativity of graph structure (LGMD$+$) in real-world applications with superior performance. Numerical experiments on real-world data show that LGMD (and LGMD$+$) is competitive or superior to the competing methods and could obtain better low-rank representations.
	
\par There are several questions remained to be investigated. First, the estimation of the precision matrix is still inefficient for very large data. The objective function involving the square root of the precision matrix makes it difficult to solve. Second, the estimation of the precision matrix may be not stable during optimization. LGMD$+$ alleviates this problem in some sense but is relatively more time-consuming.

%

	\ifCLASSOPTIONcaptionsoff
	\newpage
	\fi

	
	
	%
	\bibliographystyle{IEEEtran}
	\bibliography{IEEEabrv,reference}

%
%

\end{document}